\documentclass[sigconf]{acmart}

\usepackage{microtype}
\usepackage{graphicx}
\usepackage{subfigure}
\usepackage{booktabs} 
\usepackage{amsmath, amsfonts}
\usepackage{mathrsfs}
\usepackage{comment}
\usepackage{pgfplots}
\usepackage{tikz}
\usetikzlibrary{patterns}
\graphicspath{ {./img/} }

\tikzstyle{mystep} = [rectangle, rounded corners, minimum width=3cm,text centered, text width=3cm, draw=black]
\tikzstyle{arrow} = [thick,->,>=stealth]

\usepackage[ruled,vlined]{algorithm2e}
\usepackage{commath}
\usepackage{booktabs}
\usepackage{hyperref}
\usepackage{bbm}

\AtBeginDocument{%
  \providecommand\BibTeX{{%
    \normalfont B\kern-0.5em{\scshape i\kern-0.25em b}\kern-0.8em\TeX}}}

\copyrightyear{2021} 
\acmYear{2021} 
\setcopyright{acmlicensed}\acmConference[KDD '21]{Proceedings of the 27th ACM SIGKDD Conference on Knowledge Discovery and Data Mining}{August 14--18, 2021}{Virtual Event, Singapore}
\acmBooktitle{Proceedings of the 27th ACM SIGKDD Conference on Knowledge Discovery and Data Mining (KDD '21), August 14--18, 2021, Virtual Event, Singapore}
\acmPrice{15.00}
\acmDOI{10.1145/3447548.3467165}
\acmISBN{978-1-4503-8332-5/21/08}

\DeclareMathOperator*{\argmax}{argmax}
\DeclareMathOperator*{\argmin}{argmin}

\begin{document}

\title{Contextual Bandit Applications in a Customer Support Bot}

\author{Sandra Sajeev}
\authornote{Both authors contributed equally to this work.}
\email{ssajeev@microsoft.com}
\affiliation{%
  \institution{Microsoft Azure AI}
}

\author{Jade Huang}
\authornotemark[1]
\email{jadhuang@microsoft.com}
\affiliation{%
  \institution{Microsoft Azure AI}
}

\author{Nikos Karampatziakis}
\email{nikosk@microsoft.com}
\affiliation{%
  \institution{Microsoft Azure AI}
}

\author{Matthew Hall}
\email{mathall@microsoft.com}
\affiliation{%
  \institution{Microsoft Azure AI}
}

\author{Sebastian Kochman}
\email{sebastko@microsoft.com}
\affiliation{%
  \institution{Microsoft Azure AI}
}

\author{Weizhu Chen}
\email{wzchen@microsoft.com}
\affiliation{%
  \institution{Microsoft Azure AI}
}

\fancyhead{}
\renewcommand{\shortauthors}{Sajeev and Huang, et al.}

\begin{abstract}
Virtual support agents have grown in popularity as a way for businesses 
to provide better and more accessible customer service.
Some challenges in this domain include ambiguous user queries as well 
as changing support topics and user behavior (non-stationarity). We do, 
however, have access to partial feedback provided by the user 
(clicks, surveys, and other events) which can be leveraged to improve the user experience.
Adaptable learning techniques, like
contextual bandits, are a natural fit for this problem setting. 
In this paper, we discuss real-world implementations of contextual 
bandits (CB) for the Microsoft virtual agent.
It includes intent disambiguation based on neural-linear bandits (NLB)
and contextual recommendations based on a collection of  multi-armed bandits (MAB). 
Our solutions have been deployed to production and have improved 
key business metrics of the Microsoft virtual agent, as confirmed 
by A/B experiments. Results include a relative increase of over 12\% in 
problem resolution rate and relative decrease of over 4\% in escalations 
to a human operator. While our current use cases focus on intent 
disambiguation and contextual recommendation for support bots, 
we believe our methods can be extended to other domains.

\end{abstract}




\maketitle

\section{Introduction}
\begin{figure}[t]
\centering
\frame{\includegraphics[width=0.34\textwidth]{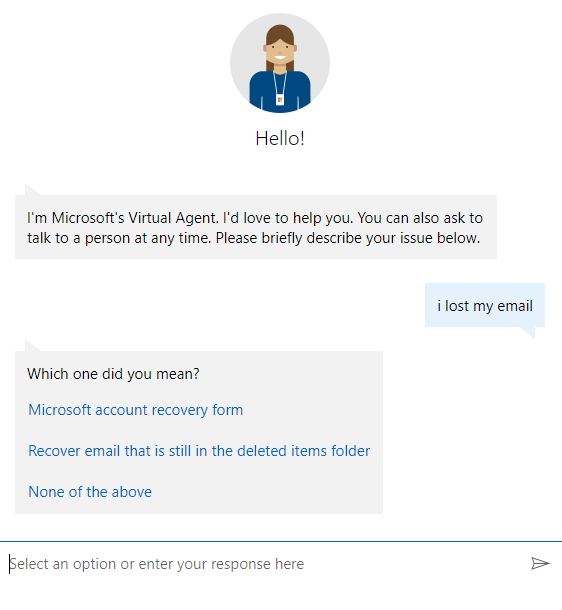}}
\caption{The Microsoft virtual agent. In response to the user query, an intent clarification dialog presents a \emph{slate} of topics generated by a policy trained using NLBs.}
\label{fig:dym}
\end{figure}

Virtual support agents have become ubiquitous in customer service for businesses. The Microsoft virtual agent is a customer support agent with which millions of users engage everyday. 
A typical session between the virtual agent and the user is shown in figure \ref{fig:dym}.
Challenges that come with operating this system include adapting to changing user behavior and support topics. For instance, bugs associated with the release of new operating system $Y$ can result in users asking more questions about $Y$ rather than previous operating system $X$. In addition, editors will create new support articles about operating system $Y$ to help address common problems; such articles should be suggested by the system when relevant as opposed to outdated content. Another example is that the rise in employees working from home due to a pandemic may result in users asking more questions than before about dual monitor detection or how to connect audio wirelessly.
Machine learning, specifically reinforcement learning, has many opportunities to help optimize virtual support agents and meet these challenges head-on. 

We drew inspiration from advances in recommendation
systems in news \cite{li2010contextual} and video \cite{schnabel2016recommendations, chen2019top}. The industry standard for recommendation systems has recently been switching to 
CBs \cite{langford2007epoch}, a simplified single-step reinforcement
learning (RL) paradigm which requires exploration but does not require dealing with credit assignment. In addition, CBs are data-efficient, which is helpful in our scenario where exploration of riskier actions can hurt the end-user experience and have negative monetary implications. In the context of the virtual agent, we deal with a partial information setting, meaning we only receive feedback (click, survey response, etc.) for the actions the user took. This can easily be modeled with CBs.

Our work presents applications of CBs for two scenarios of the Microsoft virtual agent: intent disambiguation and contextual recommendations. In intent disambiguation, the goal is to provide recommendations that tailor to the user query about a topic. 
For contextual recommendations, the objective is to provide a set of recommendations to the user as soon as they engage the virtual agent prior to any user query, based on a set of context features. These contextual features can include information such as the website from where the user engages the virtual agent.

We have deployed a MAB solution for contextual recommendations and a NLB solution for intent disambiguation, both of which showed positive impact on key performance indicators (KPIs) in online A/B experiments.

The rest of the paper is organized as follows. First, we present the problem setting and notation. Next, we describe the methods used, going into detail about the learning paradigms. Then, in the implementation section, we outline how we incorporated MABs for contextual recommendations and NLBs for intent disambiguation. Finally, we present the efficacy of our solutions through results from online experiments in the evaluation section. 

\section{Problem Setting}
The application motivating our work is the Microsoft virtual agent -- an interactive dialogue system providing first-line customer support for many Microsoft products. 
The virtual agent can be accessed via multiple channels -- most commonly through the Microsoft support website\footnote{\url{https://support.microsoft.com/en-us/contactus}} or the Get Help app which comes with Windows 10.

In the Microsoft virtual agent's original architecture, manually configured rules dictated the behavior of the system.
Most of these rules were not important from a business perspective, but rather assumptions made by application developers due to a lack of data. 
Moreover the original system could not adapt automatically to the diversity of user intents and a changing environment, such as software updates causing new issues or updated support content.
RL addresses some of these challenges and is a good fit for this problem setting for the following reasons:

\begin{enumerate}
    \item Microsoft products (including Xbox, Office, Windows, Skype) have a large customer base, hence the virtual agent's incoming traffic is also significant, making RL methods feasible. 
    \item We have access to multiple feedback signals including click behavior, 
    escalation to a human agent, and responses to survey questions such as ``Did this solve your problem?''. Some of these signals are tracked by the product team as their KPIs.
\end{enumerate}

At the same time, the application poses interesting challenges for the use of RL.
Goal-oriented dialogue systems need to not only understand the user's original intent,
but also be able to carry the state of the dialogue and guide the user towards their
goal. While our long-term plan is to be able to have a single RL agent in charge of
the whole conversation, the requirements of such an endeavor, including the of
number of samples needed to learn non-trivial policies, are substantial.  
Therefore, we started by applying RL to the virtual agent in isolated components first, ignoring issues of credit assignment and thus working in the setting of CBs.

In this work, we focus on two specific scenarios within the Microsoft virtual agent. In the following subsections, we describe them in more detail, and then formalize the notation that is shared between both scenarios.

\subsection{Intent Disambiguation}
\label{scenario_intent_disambiguation}
The domain of customer support, especially for a large company such as Microsoft, is complex and has a significant number of intents. Our intent disambiguation policy is tasked with deciding when the user's query is clear enough to directly trigger a solution, such as a troubleshooting dialogue,
or to ask a clarification question.

The inputs to this policy are the statement of the issue by the user (the query), user context features, and a list of candidate actions. The query can range from a few keywords to long and complex sentences or even paragraphs. The user context includes information like the user's operating system and its version (e.g.,\ Office products work on Windows, Mac, iOS, and Android).
The candidate actions are a collection of pre-authored dialogue intents or solutions related to the user's query pulled from the Web. Retrieval of these candidates is currently performed by several strategies. One of them includes a deep learning model similar to the one described in \cite{humeau2020poly}. Another uses Bing Custom Search for customized document retrieval. These retrieval components are currently out-of-scope for RL-based optimization so we will focus on the policy that operates with a small list of retrieved candidates.

Given the input, the policy can do one of the following:
\begin{itemize}
    \item Directly trigger a single intent or a solution: this can start a troubleshooting dialogue or display a rich text solution.
    \item Ask a yes/no question: ``Here's what I think you are asking about: \ldots Is that correct?''
    \item Ask a multiple-choice question: ``Which one did you mean?'', followed by titles of two to four intents as well as option ``None of the above.''
    \item Give up: ``I'm sorry, I didn't understand. It helps me when you name the product and briefly describe the issue.'' 
\end{itemize}
Figure \ref{fig:dym} presents an example of a multiple-choice question action taken by the disambiguation policy. 

\subsection{Contextual Recommendations}
\label{scenario_contextual_recommendations}

The ``Settings'' app is a desktop application in Windows 10, where a user can click ``Get help'' from any Settings page (e.g.,\ Bluetooth, Display etc.) and interact with the Microsoft virtual agent. We will refer to the page the user is coming from as the \emph{source page} - it is a crucial part of the user context that is available to us.

\begin{figure}[t]
\centering
\frame{\includegraphics[width=0.34\textwidth]{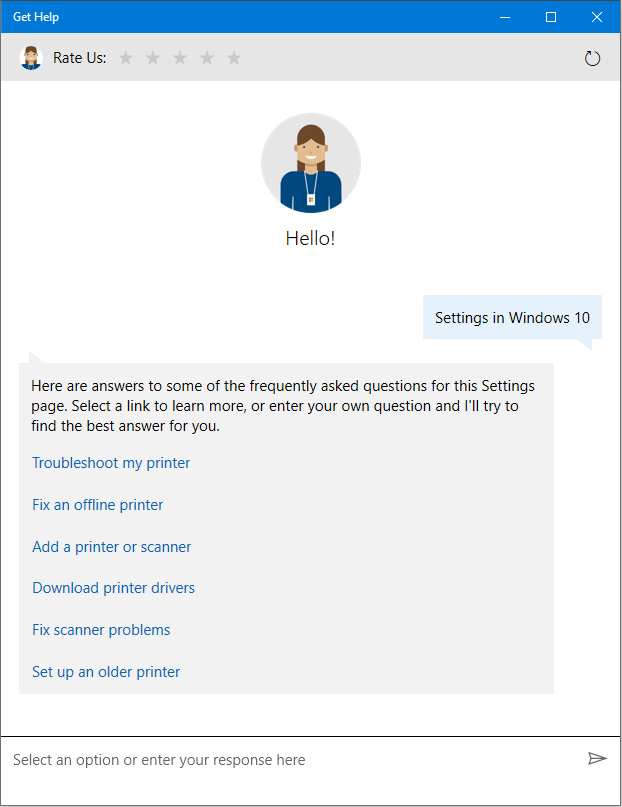}}
\caption{Contextual recommendations in Settings in Windows 10, for the source page ``Printers''.}
\label{fig:siw10}
\end{figure}

The goal is to provide contextual recommendations, i.e., to recommend a \emph{slate} (sequence) of solution topics before the user types anything, simply based on the context sent by the app (see figure \ref{fig:siw10}).
The baseline experience is a fixed mapping from source pages to slates of up to six topics manually picked by human editors. 

We were motivated by several reasons to leverage feedback data to learn better contextual recommendations and improve upon the baseline. First, maintaining manually-specified dialogues does not scale to the approximately 200 source pages present in the app 
(most of them served the same fixed slate of six topics). Second, a large amount of traffic flows through the app, which suggests that data-driven techniques could do well here.
Third, the available user context includes more signals than just source page, which can help with suggesting relevant solutions in certain situations, such as the device network type (wired, wifi) and battery status (charging, discharging).

Given the context, the agent can take one of the following actions:
\begin{enumerate}
    \item Suggest up to six solutions out of about 3000 candidates. 
    \item Choose to fall back to default behavior of recommending six fixed options chosen by editors for the current source page.
\end{enumerate}

\subsection{Feedback signals and goals}
\label{feedback_signals}
The feedback signals available to learn from are the same in both scenarios described above:
\begin{itemize}
    \item \textbf{Click.}  When a slate of topics is presented, the option selected by the user is recorded. This signal is censored in scenario \ref{scenario_intent_disambiguation} when the policy decides to directly trigger.
    While useful, it is not a metric important to the business, so we do not optimize for it alone.
    \item \textbf{Survey.}  After providing a final answer to the user, the virtual agent asks whether the solution has resolved their problem. The user may respond ``yes'', ``no'', or decline to answer. The product team tracks this signal, aggregated per user session and averaged, as the most important KPI called Problem Resolution Rate (more details in section \ref{metrics}).
    Hence, this feedback signal is also our main reward signal.
    \item \textbf{Escalation.} The user can decide at any time to talk to a human agent. This negative reward signal is directly related to the actual cost of running a call center. In general, this metric 
    is monitored but is not being optimized.
\end{itemize}

\subsection{Notation}
\label{notation}

The contextual recommendation and intent disambiguation scenarios that we focus on in this paper can be modeled as combinatorial CBs.
We adapt and expand notation proposed in  \cite{swaminathan2017offpolicy} to the scenarios found in the Microsoft virtual agent.

In this setting, an agent interacts with the environment repeatedly as follows:

For time step $t = 1, 2, \ldots$: 
\begin{enumerate}
    \item The world produces a \emph{context} $x_t \in \mathcal{X}$ (e.g., the user sends a query $x_t$ to the system).  
    \item The agent chooses a slate $\mathbf{s_t} = (a_1, ..., a_l) \in \mathcal{S}(x_t)$,
    of variable length $l$
    ($0 \le l \le L$),
    consisting of actions $a_j \in \mathcal{A}(x_t)$.
    We call $\mathcal{A}(x_t)$ an \emph{item action space} and $\mathcal{S}(x_t)$ a \emph{slate space}.
    Position $j$ in a slate is called a \emph{slot}.
    \item Given the context and slate, the world produces \emph{feedback signals},
    including:
    
    \begin{itemize}
        \item $click_t \in \{0,1\}^l$ informing of the slot selected by the user. In this work, we assume it is either a one-hot vector or a zero vector (i.e., no more than a single item can be selected).
        \item $survey_t \in \{\mathrm{yes}, \mathrm{no}, \mathrm{skipped}\}$ indicating the user's assessment of the solution presented by the virtual agent, provided via a survey. Values ``yes'' and ``no'' mean positive and negative feedback, respectively, while ``skipped'' means the user has not responded to the survey.
        \item $escalation_t \in \{0, 1\}$ indicates whether the user has decided to escalate the case to a human agent ($1$) or not ($0$).
    \end{itemize}
    
    \item The reward for each slot $j$ in the slate $\mathbf{s}_t$ is denoted by $r_{t,j}$. It is computed as a function of one or more of the \emph{feedback signals}.
\end{enumerate}

Note that the item action space $\mathcal{A}(x_t)$ is not a fixed set but can vary based on the context.
The actions are parametric, with features such as title, type of content, etc. In our case, the majority of actions represent support documents and troubleshooting dialogues, but $\mathcal{A}(x_t)$ may also include the special action \emph{null item}, denoted as $\bot$. It gives the user a choice of indicating that none of the other actions is relevant. If selected via $click_t$, the system moves on to other suggestions.

To gain intuition about this notation, let's consider some examples:

\begin{itemize}
    \item \textbf{The intent disambiguation scenario presented in the figure \ref{fig:dym}}. The query ``i lost my email'' is the main property of the context $x$. Topics ``Microsoft account recovery form'' and ``Recover email that is still in the deleted items folder'' are some actions $a_1$ and $a_2$. The ``None of the above'' option rendered in the dialog translates to the \emph{null item} $\bot$ - it gives the user an opportunity to explore other topics or ask another question. Hence, the slate chosen by the policy at that time step was $\mathbf{s} = (a_1, a_2, \bot)$.
    
    The intent disambiguation policy is also allowed, for some contexts, to directly trigger a topic. Such a slate would look like $\mathbf{s} = (a_1)$, without the $\bot$ action. What kind of slates are allowed in what contexts depends on the function $\mathcal{S}(x)$.
    
    \item \textbf{The contextual recommendations scenario presented in the figure \ref{fig:siw10}}.  The source page ``Printers'' is the main property of the context $x$. Ignoring the presented topics and typing a query manually is equivalent to selecting the \emph{null item} $\bot$ - hence, the slate can be represented as $\mathbf{s} = (a_1, a_2, \ldots, a_6, \bot)$.
    
    In the contextual recommendations scenario, every slate in $\mathcal{S}(x)$, in every context $x$, must always include $\bot$ (i.e.: topics are never directly triggered).
\end{itemize}

The reward used by our CB algorithms is always a function of feedback signals.
Decoupling them from the reward definition, while not a standard approach in the RL literature (as in \cite{langford2007epoch, chapelle2011an}), is useful in the learning algorithms we are presenting in later sections. 

Some feedback signals may be delayed. For example, a long conversation may occur between the action $a_t$ and the point at which we ask the user if the
proposed solution addressed their problem (to generate the signal $survey_t$). We currently treat all variations that can happen there 
as part of the environment that generates a noisy version of the reward. This simplification is a practical choice and can be removed with better
modeling of the interaction between the user and the bot. 

\section{Methods}
We divide discussion about the solutions into two main categories: learning from a discrete context (corresponding to the contextual recommendations scenario described in section \ref{scenario_contextual_recommendations}) and learning from a rich context (relevant in the intent disambiguation scenario from section \ref{scenario_intent_disambiguation}).

\subsection{Learning with Discrete Context}
For each discrete context $x_t \in \mathcal{X}$, let there be two MABs corresponding to two feedback signals: clicks and survey responses. At time step $t$, each MAB can be described as follows:

\begin{itemize}
    \item We have collected a history of reward successes and failures over $w_x$ total past time steps in each context $x$ separately for each action $a$. This $w_x$ is specific to each context. 
    \item We have a function $Sample(successes, failures)$ that samples scores for all actions.
    \item Actions are ordered by these sampled scores.
    \item The slate ends at the position at which $\bot$ is ranked.
\end{itemize}

For the click bandit and context $x$, we consider
\begin{itemize}
    \item successes: the total instances action $a$ has been clicked for the past $w_x$ time steps
    \item failures: the total instances an action $a$ has been observed (ranked above $\bot$) but not clicked
\end{itemize}

For the survey response bandit and context $x$, we ignore the cases that the survey is not answered, considering
\begin{itemize}
    \item successes: total ``yes'' answers
    \item failures: total ``no'' answers
\end{itemize}

\subsubsection{Sampling with a Discrete Context}
\label{sample-discrete}
To balance exploitation with exploration, we use a sampling algorithm to produce plausible
estimates of the probability of a click and the probability of a ``yes'' survey. We compared both Thompson sampling \cite{thompson1933, russo2020tutorial} and EwS~\cite{maillard2011apprentissage}, and qualitatively we found that results in our application looked better with Thompson sampling so we focus on it here. However, EwS is reasonable to use as well.

Let $\alpha_t^{x,a}$ represent the successes for action $a$ in context $x$, given the data until time step $t$, within window $w_x$. Likewise, let $\beta_t^{x,a}$ represent the failures for action $a$ in context $x$, given the data until $t$, within window $w_x$.
We start with a uniform prior distribution 
$Beta(\alpha=1, \beta=1)$ and assume each event 
is iid. Then the posterior probability is $Beta(\alpha_t^{x,a}+1, \beta_t^{x,a}+1)$. 
In other words, we add one to all success and failure counts. In practice, we track successes $\alpha$ and trials $N$, and failures are then computed as $\beta = N - \alpha$.

Each bandit samples an estimated action value $Q(x_t, a)$ from the posterior distribution $Beta(\alpha_t^{x_t,a}+1, \beta_t^{x_t,a}+1)$ for each action $a$. To represent the final score for an action $a$ based on both bandits, we initially combined the scores from the two bandits using a naive product to model the joint probability $P(click) \cdot P(survey=yes$|$click)$.  Later, having amassed more logged data via online experimentation, we switched to a more flexible approach of using a context-specific interpolation score $\lambda$ which weights the click score lower if there are sufficient survey response trials.

\begin{equation}
\label{eq:backoff}
\log(Q_{joint}(x_t,a)) \triangleq \lambda_x \log(Q_{click}(x_t,a)) + (1 - \lambda_x) \log(Q_{survey}(x_t,a))
\end{equation} 

This combined score is used to rank actions in descending order. The size of the slate visible to the user is determined by the minimum of the position of $\bot$ in the slate and action space rules dictating the maximum slate length. In other words, the slate that the user actually sees is a subset of the scored actions. All actions with a higher sampled score than $\bot$ is considered observed by the user. 

Having assembled the slate, the user then provides feedback signals. As mentioned in \ref{notation}, the $click$ signal is a one-hot vector: 1 for the clicked slot and 0 for the rest. The $survey$ signal is considered observed only for the slot that was clicked.

Note that the downside of Thompson sampling in comparison to EwS is that it does not produce closed-form probabilities of choosing an action in a given state. Collecting such probabilities is often useful for off-policy learning and evaluation methods. This can be mitigated by logging the $\alpha$ and $\beta$ values at the time of making a decision and estimating these probabilities offline in a Monte-Carlo fashion.

\subsection{Learning with Rich Context}
\label{learning-in-rc}
A rich context is defined by large cardinality of the set $\mathcal{X}$, which makes it difficult to learn the value function for each state. Therefore value function approximation is required for rich contexts, like natural language embeddings and image data. 
Assume there is a representation function $\phi$ that takes in a rich context $x \in \mathcal{X}$ 
and action $a \in \mathcal{A}$ and outputs a set of low-dimensional features, $\phi(x,a) \in \mathbb{R}^{d}$. 
We assume that the expected reward is linear with respect to these features.
\begin{equation}
\label{eq:equation1}
    \mathbb{E}[r|x, a] = w^\top \phi(x,a)
\end{equation}

In other words, there exists an unknown vector $w$, which models the linear relationship of the expected reward $\mathbb{E}[r|x,a]$, for action $a$ and a context $x$. 

\subsubsection{Neural-Linear Bandit}
\label{nlb}
Correct quantification of 
uncertainty is critical in bandit algorithms because the algorithm is in charge of how data is collected. If the algorithm uses  
uncertainty estimates that are too wide, it over-explores and unnecessarily tries 
suboptimal actions. If the uncertainty estimates are too narrow, the algorithm
under-explores and runs the risk of not discovering the best action.

In this work we use the Neural-Linear Bandit strategy
from \cite{riquelme2018deep}, 
which compares various methods for quantifying uncertainty with deep learning models. There it was demonstrated that the neural-linear approach is often the best single-model method (i.e., not relying on an ensemble) for uncertainty quantification. We have slightly adapted this method for our purposes.
The NLB consists of two main parts. First is the representation function $\phi: \mathcal{X} \times \mathcal{A} \mapsto \mathbb{R}^{d}$ which produces a \textit{d}-dimensional vector of neural features.
The function $\phi$ is generated by fitting a neural network, $N: \mathcal{X} \times \mathcal{A} \mapsto \mathbb{R}^{1}$, which predicts the observed reward from query-answer pairs.
The last layer of this network is removed to represent $\phi(x, a)$.

The second part is the bandit function, where we try to estimate the $w$ that satisfies \eqref{eq:equation1} and the uncertainty around it. This is the simplifying assumption in NLBs: all uncertainty in the model can be approximated just by the uncertainty in the last layer of the network. Due to the assumption that the reward is linear with respect to the neural features, the focus is on creating a posterior distribution for the solution. 
Note that validating this assumption becomes difficult in practice due to the large dimensions of the neural features. 
\begin{equation}
    \label{eq:least-squares}
    \hat{w}_t = \argmin_w \sum_{i=1}^{t-1} (\phi(x_i,a_i)^\top w - r_i)^2
\end{equation}
This is a simple least squares problem that we need to solve each time we 
update the bandit. To do this efficiently and support rolling windows 
of training data we maintain the sufficient statistics:
\begin{align}
    f_t &= \sum_{i=1}^{t-1} r_i\phi(x_i, a_i)\\
    B_t &= \sum_{i=1}^{t-1} \phi(x_i, a_i)\phi(x_i, a_i)^\top
\end{align}

which help us rewrite eq.~\eqref{eq:least-squares} as $B_t \hat{w}_t = f_t$. 
We solve the latter with the help of the Cholesky factorization $L_t L_t^\top = B_t$
which we also use for obtaining the uncertainty for new predictions $\hat{r}(x,a)=\phi(x,a)^\top \hat{w}_t$. To do this we need a notion
of a count of how many times action $a$ has been
tried in context $x$. For linear bandits this is given by
\begin{equation}
\label{eq:bonus}
    b(a) = \frac{1}{\phi(x,a)^\top B_t^{-1} \phi(x,a)} = \norm{ L^{-1}\phi(x,a) }^{-2}
\end{equation}
The bandit gives more benefit of doubt to an action $a$ with small $b(a)$.
Similarly to the number of times an action has been tried in the case 
of a MAB, $b(a)$ is unit-less even if $\phi(x,a)$ has units:
the matrix $B_t^{-1}$ cancels them. The inverse of the quadratic
form is also necessary for other reasons. If $\phi$ were one-hot vectors, then
$B_t$ would be diagonal with the count for action $a$ in context $x$ on 
the diagonal. Therefore, the inverse quadratic form is the count.

\subsubsection{Sampling with a Rich Context}
The NLB policy samples actions $a$ to generate the slate $s$ according to a sampling algorithm. 
The main algorithms we explored for sampling in the rich context are the linear versions of Thompson sampling \cite{agrawal2013thompson} and EwS \cite{abbasi2013online}. Both sampling algorithms result in stochastic policies. 

Linear Thompson sampling \cite{agrawal2013thompson} requires specification of a prior covariance matrix for $w$, typically chosen to be a multiple of the identity matrix.
However, there is no easy way to set this hyperparameter as it can have
a hard-to-predict effect on the learning dynamics.
In addition, this algorithm doesn't provide the probability of sampling action $a$ in an explicit form which is useful for off-policy evaluation. 

Linear EwS~\cite{abbasi2013online} 
was chosen since it has no requirement of setting a prior.
In Linear EwS, the probability of selecting an action $a$ has an 
explicit form. Given context $x$ we first compute the action 
the model currently prefers 
\[
a^* = \argmax_{a \in \mathcal{A}} \hat{w}^\top \phi(x,a)
\]
and the gaps between the best action and all actions
\begin{equation}
    g(a) =  \hat{w}^\top\phi(x,a^*) - \hat{w}^\top\phi(x,a)
\end{equation}
Finally, as noted in \cite{abbasi2013online}, Linear EwS samples action $a$ proportional to $\exp(-2b(a)g(a)^2)$ where $b(a)$ is given by~\eqref{eq:bonus}. 

\subsection{Inference Strategies} 

\subsubsection{Making Exploration Safer}
\label{safer_exploration}
If a strong baseline already exists, to make exploration safer, we can compare the total value of the sampled slate to the value of the baseline slate. Whichever slate has a higher value is then presented to the user. In our case, we consider the slate value to be the sum of the sampled scores of the actions in the slate. This simple trick can reduce variance in the KPIs, normally expected from exploration, after initial deployment of an automatic learning system. Note that the baseline slate can be still used to collect feedback signals and improve future predictions.

\subsubsection{Sampling to Obtain a Slate}
\label{sample-once}
One can simply sample for expected rewards from possible actions once to obtain a full slate of actions. If there are per-slot rules dictated by business requirements, they can be applied after the sampling.

\section{Implementation}

This section outlines key implementation details about our MAB solution for contextual recommendations and NLBs for intent disambiguation. Both training pipelines were written in Python. All training and evaluation was done asynchronously using Azure Batch. The inference code was written in C\# to easily integrate with our partner team's product, the Microsoft virtual agent. The intent disambiguation models were trained using PyTorch \cite{NEURIPS2019_9015} and converted to ONNX \cite{bai2019} for inference in the production system. ONNX Runtime is used to load the models in .NET.

\subsection{Contextual Recommendations}
\label{contextual-recommendations}

\subsubsection{Problem Formulation}

The discrete context in our case is categorical: the source page from where the user is clicking ``Get help''. In offline analysis, we found that while there are other interesting attributes in the context such as battery value and network type, the source page is the most indicative of what the user may free-type when they choose not to select from the given list of ``top solutions''. For example, users clicking ``Get help'' from the Printers page tend to ask printer-related questions. The candidate content is represented simply by a unique id and no other features. We associate the null item $\bot$ with the user free-typing. 

User clicks and surveys are logged and aggregated every four hours using asynchronous jobs over a moving window. This moving window ensures that content that is not performing well and not interacted with will disappear from recommendations eventually. Note that this window is context-specific, as content performing poorly in one context does not have an effect on another context.

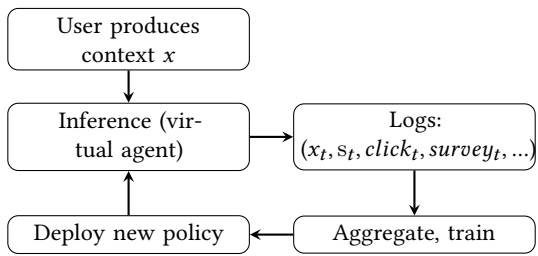
\begin{figure}[t]
    \centering
    \begin{tikzpicture}[node distance=1.3cm]
    \node (user) [mystep] {User produces context $x$};
    \node (va) [mystep, below of=user] {Inference (virtual agent)};
    \node (logs) [mystep, right of=va, xshift=2.5cm] {Logs: ($x_t, \mathbf{s}_t, click_t, survey_t, ...$)};
    \node (deploy) [mystep, below of=va] {Deploy new policy};
    \node (aggr) [mystep, below of=logs] {Aggregate, train};

    \draw [arrow] (user) -- (va);
    \draw [arrow] (va) -- (logs);
    \draw [arrow] (logs) -- (aggr);
    \draw [arrow] (aggr) -- (deploy);
    \draw [arrow] (deploy) -- (va);
    \end{tikzpicture}
    \caption{Feedback loop for contextual recommendations.}
    \label{fig:loop}
\end{figure}

At runtime, a user clicks on ``Get help'' from source page $x$. First, we gather available candidates that have click and survey counts. We have limitations on the size of our logged data, so we perform an initial sampling of the candidates from over 1000 candidates down to around 25 using Thompson sampling as described in \ref{sample-discrete}. 

With this pared-down set of candidates, we sample once more using Thompson sampling to obtain a final slate of actions. The length of the slate is determined by the minimum of the following: the null item's position and the configured max length of the slate $L$ (in our case $L=7$ including $\bot$). We apply the trick of making exploration safer from \ref{safer_exploration} as there is a strong human-authored baseline.

\subsubsection{Automatic Action Space Expansion}
So far as we've described, this experience has a concrete set of candidates per source page from which it draws to recommend to the user. Occasionally, an editor may author new content relevant to a source page which performs well within the context of a source page but outside of the initial recommendation, e.g., the user free-types a query that triggers this content, finds it helpful, and answers `yes' to the end-of-chat survey. Thanks to our work with intent disambiguation, these interactions are also logged.

To address this, we have another asynchronous job that runs once a day. This job merges statistics from our intent disambiguation flow that occur within the context of a source page with statistics from the initial Settings recommendations flow. For each source page, we use the null item's score as a success threshold for candidates in consideration from intent disambiguation. We first filter out candidates with insufficient trials, e.g., 10 trials. For each potential candidate $a$ we then compute
a bound on the probability that the expected reward of $a$ is larger
than the expected reward of $\bot$. 

We then filter out candidates whose bound on the probability is less than our configured allowable false positive probability $fp$. Intuitively, this allowable false positive probability is the maximum probability we allow that an individual new candidate has lower survey response success than $\bot$, i.e., the probability of a candidate not being a false positive is $\geq 1-fp$. The qualifying candidates' counts are copied over from the intent disambiguation space to the Settings space--however, we zero out the click counts as the trials between the two spaces are not comparable.

\subsection{Intent Disambiguation}
\label{intent-disambiguation}

\subsubsection{Problem Formulation}
For the intent disambiguation scenario, the context is rich. 
The input data to the policy contains the text embedding of the user query and candidate title generated in a manner similar to \cite{humeau2020poly}. 
It also includes a collection of categorical user context features, such as the website from where the user initiated the virtual agent. When the user engages the virtual agent by typing a question, our policy samples from a set of candidates described in \ref{sample-once} to generate a slate. The slate length is determined by the null item.

\subsubsection{Training}

To train the NLB policy, we first featurize the logged data, limiting to sessions that have explicit user survey responses (``yes'' or ``no''). We initially defined the reward to be a function of the survey feedback of only the clicked action \eqref{eq:reward-def-nlb}. 
\begin{equation}
\label{eq:reward-def-nlb}
  r_t = \begin{cases} \mbox{1,} & \mbox{$survey_t$ = ``yes''}   \\
 \mbox{-1,} &  \mbox{$survey_t$ = ``no''} \\ \end{cases} 
\end{equation}
In subsequent iterations, we experimented with combining multiple feedback signals for the reward as described in section \ref{nlb-experiments}.
After featurization, we follow the learning procedure for NLBs defined in \ref{nlb}. 
The training objective of the network $N$ was set to minimize the squared error between the reward of the clicked action and the prediction. 
This network is used to generate $\phi$ with $d=2048$ neural features. 
Note that $d$ cannot be excessively large as the bandit needs to store 
$B_t^{-1}$ or $L_t^{-1}$ which is of size $d \times d$ to compute 
eq.~\eqref{eq:bonus}. 

We train the bandit in fixed periods (e.g., one day). 
For each fixed period, we solve the least squares problem defined in \eqref{eq:least-squares}. 
When training this model in practice, the matrix $B_t$ can be very close to low rank. 
To be able to have a usable inverse, we used principal 
component regression and retained enough principal components to capture
99\% of the variation in the neural features.

The initial NLB policies were fixed: trained on a set number of days of data and deployed to production. We later explored auto-updating policies with hourly retraining.

\section{Experiment results}
\label{experiments}

\subsection{Metrics}
\label{metrics}

The virtual agent product team tracks several KPIs where some were the metrics for which we optimized. 
Our metrics are reported over user sessions and each user session is treated independently. We do not track an individual user's behavior over time, e.g. if a user re-engages with the Microsoft virtual agent, that is treated as a separate user session.

\begin{itemize}
\item \textbf{Problem Resolution Rate (PRR):} a ratio of user sessions with a positive survey response to all sessions with any survey response. This is the main metric we track while running offline or online experiments.
\item \textbf{Engaged Assisted Support (EAS):} fraction of user sessions that end up with escalation to human-assisted support.
\item \textbf{Self Help Success (SHS):} fraction of user sessions that ended with assumed success (i.e., an answer is delivered and the user does not engage assisted support or indicate the answer was not useful).
\item \textbf{User Engagement (UE):} fraction of user interactions where the user sends at least one message after the virtual agent's greeting (where the greeting may include a slate with contextual recommendations, in case of help for the Settings app).
\end{itemize}

We can approximate some of these metrics using feedback signals described in sections \ref{feedback_signals} and \ref{notation}. E.g., for a time period $t = 1, ..., T$ we can define the following approximations:

\begin{eqnarray*}
\hat{PRR}_{1:T} &=& \frac{\sum_{t=1}^T \mathbbm{1}\left[ survey_{t}=yes\right]}{\sum_{t=1}^T \mathbbm{1}\left[ survey_{t}=yes \vee survey_{t}=no\right]} \\
\hat{EAS}_{1:T} &=& \frac{1}{T}\sum_{t=1}^T  escalation_{t}
\end{eqnarray*}

Where $\mathbbm{1}$ is an indicator function.

The difference between the product team's KPIs and the approximated values is that the real metrics are aggregated across a user session rather than across an individual event within a user session. A single session lasts from the moment the user connects with the virtual agent until the last event that occurs within the conversation. Specific aggregation depends on the metric - e.g., the user may respond negatively to one survey, but then continue conversation with the virtual agent, and eventually respond to the second survey positively. The product team's KPI would indicate this as a single event with success contributing towards PRR, while our simplified definition would consider this as two events - one failure and one success.

Our CB algorithms, as currently defined, are capable of optimizing only the approximated metrics like $\hat{PRR}$ or $\hat{EAS}$. While aware of this problem, we observe that a CB policy optimizing a simplified metric will usually cause movement of the real KPI in the same direction, when deployed to production, as the following sections show.

\subsection{Offline Evaluation}
\label{offline-eval}
Before attempting online experimentation with the NLB policy for intent disambiguation, we first used offline evaluation to check whether a newly trained policy seemed promising. This helped limit the negative impact to actual users in production traffic. Since we made use of EwS sampling, our behavior policy was stochastic and we logged the probabilities. 
This allows us to run off-policy evaluations to estimate the impact of our policy.

We use the SNIPS \cite{swaminathan2015the} estimator to measure how much more reward our current policy would have provided compared to the logging policy. 
In our counterfactual evaluation, we were only able to use data where the logged policy and the new policy agreed on the clicked action, rather than the entire slate. 
The primary metrics we observed were how the logged policy and new policy compared in terms of rewards based on the survey and deflection feedback signals.
A policy had to achieve higher reward estimates than the logging policy with low variance to be promoted to an online experiment.
The requirement of low variance was enacted to reduce mismatch between offline evaluation and online experiments. 

In the case of contextual recommendations, we do not perform offline evaluation beyond qualitative analysis as described in \ref{sample-discrete} and careful monitoring of the auto-updated results. The updated model is deployed automatically without any experiments or human in the loop, although with an automated monitoring system alerting about unusual behavior. 

\subsection{A/B experimentation}
We conducted several online A/B experiments to verify the efficacy of both our new contextual recommendation and intent disambiguation policies in comparison to a control policy (i.e., the current deployed production policy) on user traffic from the Microsoft virtual agent. 
Each A/B experiment should be viewed independently, as the control policy and the amount of user traffic varies from experiment to experiment.
The A/B experiments gave key insights into how well the polices improved KPIs relative to control. Some of these KPIs were only approximated by offline evaluation and others were not captured in offline evaluation, like UE or SHS (see section \ref{metrics}). 
Furthermore, A/B experiments reflected the actual environment in which the policy was deployed, which differed in some cases from the training data. 
For instance, the NLB policies were trained only with data that included the survey signals, "yes" or "no", leaving out data where the user did not answer the survey.
Finally, these A/B experiments allowed us to quantify impact, which was especially key for contextual recommendations which did not have formal offline evaluation. 
The results from these A/B experiments were the deciding factor in whether to promote these new policies to the default experience, thus serving general production traffic in the future.
When looking at results from online A/B experiments, there were several KPIs that we observed as described in section \ref{metrics}. 

Optimizing for several KPIs was a balancing act. Some of these metrics like PRR were almost directly optimized for as reward signals during training, others like EAS were used as limitations where we sought to prevent harmful movement, and others such as SHS and UE were assumed to be correlated to reward signals.

\subsection{Contextual Recommendations}
We first ran an A/B experiment for an automatically updating implementation of our MAB approach based on Thompson sampling that used the naive combined score $P(click) \cdot P(survey=yes$|$click)$
described in section \ref{sample-discrete}. The control was a human-authored fixed mapping from source pages to slate. The experiment showed gains in SHS and UE, which led us to deploy this new policy to all production traffic.

Afterwards, we set up a small experiment to monitor the auto-updating policy's performance over time. The MAB policy was set as the champion policy and the old control behavior as the challenger. The challenger only ran with a small amount of traffic as to provide minimal impact on performance. Running this monitoring experiment was a crucial decision, as over time the champion policy began to show negative movement in PRR as well as smaller gains in SHS and UE. As PRR is our primary KPI, gains in SHS and UE was not worth a drop in PRR.

Motivated in part by the negative results of the monitoring experiment, we worked on implementing a more dynamic interpolation scoring approach described by equation \ref{eq:backoff}. We ran an A/B experiment for a second automatically updating MAB based on this approach. This experiment also showed positive movement in SHS and UE for the treatment compared to the control without any drop in PRR, as shown in table \ref{tab:thompson}.

\begin{table}[h]
  \begin{center}
    \caption{Dynamic Thompson Sampling Policy A/B Results}
    \label{tab:thompson}
    \begin{tabular}{llll}
      \toprule %
      \textbf{Metric} & \textbf{Movement} & \textbf{P-Value} & \textbf{Sample Size}\\
      \midrule %
      PRR & +1.13\% & 0.2384 & 36K \\
      EAS & -1.73\% & 0.3656 & 193K \\
      SHS & +2.93\% & $<10^{-10}$ & 193K \\
      UE & +2.13\% & $<10^{-10}$ & 272K \\
      \bottomrule %
    \end{tabular}
  \end{center}
\end{table}

This second policy was then deployed to production, replacing the previously deployed MAB policy. Afterwards, we set up another small experiment to monitor the new auto-updating policy's performance over time. Over several weeks, the policy began to show improvements in PRR and EAS as well as greater gains in SHS and UE as seen in table \ref{tab:thompsonreverse}. It is important to note that a reduction in EAS is a positive improvement, as it is reducing the number of escalations to a human agent and practically, reducing cost. 

\begin{table}[h]
  \begin{center}
    \caption{Dynamic Thompson Sampling Policy Monitoring A/B Results}
    \label{tab:thompsonreverse}
    \begin{tabular}{llll}
      \toprule %
      \textbf{Metric} & \textbf{Movement} & \textbf{P-Value} & \textbf{Sample Size}\\
      \midrule %
      PRR & +2.36\% & $5 \cdot 10^{-5}$ & 104k \\
      EAS & -4.09\% & $2 \cdot 10^{-6}$ & 549K \\
      SHS & +4.70\% & $<10^{-10}$ & 549K \\
      UE & +4.54\% & $<10^{-10}$ & 766k \\
      \bottomrule %
    \end{tabular}
  \end{center}
\end{table}

\subsection{Intent Disambiguation}
\label{nlb-experiments}
In the first A/B experiment we ran with our NLB policy, our initial NLB policy showed improvement over the then-current policy, a deep network trained via policy gradient, where both policies had the same input. The greatest improvement was in PRR, which showed a large increase over our previous policy, as shown in table \ref{tab:nlb}. There was also a small improvement in SHS.
\begin{table}[H]
  \begin{center}
    \caption{Neural-Linear Bandit Policy A/B Results}
    \label{tab:nlb}
    \begin{tabular}{llll}
      \toprule %
      \textbf{Metric} & \textbf{Movement} & \textbf{P-Value} & \textbf{Sample Size}\\
      \midrule %
      PRR & +12.45\% & $<10^{-10}$ & 94K \\
      EAS & -1.12\% & 0.0599 & 473K \\
      SHS & +0.49\% & 0.0297 & 473K \\
      UE & +0.07\% & 0.5374 & 632K \\
      \bottomrule %
    \end{tabular}
  \end{center}
\end{table}
We also ran an A/B experiment that compared our NLB policy with a greedy rules-based policy. This rules-based policy was the original solution for intent disambiguation in the Microsoft virtual agent. The experiment results showed that while the NLB policy improved PRR, the greedy rules-based policy had a better EAS score. This is likely because the NLB policy can determine if there is no candidate that matches the user's intent, and thus shows no result more frequently, leading to escalations. 

After promoting the new NLB policy to be the default policy for production traffic, we experimented with reducing the intent disambiguation slate size from four to three. 
The objective of reducing slate size was to decrease the likelihood of a user getting distracted by suboptimal candidates in the slate. This change was tested in an A/B experiment against a control of the NLB policy with a slate size of four. The experiment results showed very little movement amongst all KPIs. This gave us the confidence to reduce our slate size to three for all production traffic, also helping reduce the action space of our intent disambiguation policy.

To study the effect of optimizing different feedback signals, we ran an experiment with a model optimized for EAS in addition to PRR with the hope of lowering EAS. This A/B experiment shows promising results with improvements in both EAS and PRR as seen in table \ref{tab:nlb-escalation}. Interestingly, there was also a drop in SHS. We plan to pursue this method of training in the future.

\begin{table}[H]
  \begin{center}
    \caption{Neural-Linear Bandit Policy Optimized for Escalation A/B Results}
    \label{tab:nlb-escalation}
    \begin{tabular}{llll}
      \toprule %
      \textbf{Metric} & \textbf{Movement} & \textbf{P-Value} & \textbf{Sample Size}\\
      \midrule %
      PRR & +4.66\% & $<10^{-10}$ & 42K \\
      EAS & -2.29\% & 0.0160 & 263K \\
      SHS & -1.72\% & $6 \cdot 10^{-7}$ & 263K \\
      UE & -0.01\% & 0.9295 & 363K \\
      \bottomrule %
    \end{tabular}
  \end{center}
\end{table}

To achieve the objective of adapting to changing user traffic, we began experimenting with auto-updating policies. 
For this experiment, the control was the NLB policy with a slate size of three. 
The treatment auto-updating NLB policy was initialized with the same parameters as the control aside from a higher exploration rate.
The higher exploration rate was chosen to counteract the overfitting effects of training with less data. 
Then, we retrained NLB's bandit-layer hourly on new data from the treatment traffic. 
There were promising improvements in PRR over the two-week experiment.
With this experiment, we successfully closed the loop for the intent disambiguation scenario using the auto-updating NLB policy.

\begin{table}[H]
  \begin{center}
    \caption{Neural-Linear Auto-Updating A/B Results}
    \label{tab:nlb-escalation}
    \begin{tabular}{llll}
      \toprule %
      \textbf{Metric} & \textbf{Movement} & \textbf{P-Value} & \textbf{Sample Size}\\
      \midrule %
      PRR & +1.48\% & $0.0307$ & 90K \\
      EAS & -0.5\% & 0.4356 & 590K \\
      SHS & -0.0014\% & $0.1450$ & 590K \\
      UE & +0.13\% & 0.1936 & 795K \\
      \bottomrule %
    \end{tabular}
  \end{center}
\end{table}

\section{Related Work} 
\label{sec:rel}

In recent years, there have been many advances in using RL for recommendation systems. 
Policy gradient-based methods are one such approach to model recommendation systems, such as REINFORCE \cite{chen2019top}. 
This method is more stable in regards to policy convergence compared to Q-learning techniques.

Recent work on slate recommendation like SlateQ \cite{slateq} allows the decomposition of a slate's total value as a function of the items.
In the SlateQ approach, a Q-Network is trained
with a user choice model that approximates the click-through rate. 
One caveat is that SlateQ makes significant assumptions about user behavior through the use of a user choice model. 

CBs are a lightweight single-step RL method that works well with interactive systems. It is the augmented form of a K-armed bandit problem with context $x$ as formalized in \cite{langford2007epoch}.
In addition, CBs offer key properties that simplify the problem space.
For instance, the reward is only associated with the selected action. CBs enable exploration via sampling algorithms like Thompson sampling, epsilon-greedy, or EwS \cite{thompson1933, langford2007epoch, maillard2011apprentissage}.
Furthermore, CBs have been successfully utilized in recommendation systems.
In \cite{li2010contextual}, MABs modeled personalized news article recommendation with LinUCB sampling. 
CBs have also been used in search engines for ranking recommendations \cite{swaminathan2017offpolicy}.
 
Inspired by the application of CBs in traditional recommendation systems, we extend these ideas to be applied in intent disambiguation and contextual recommendations for a virtual support agent. For the implementation of our CB solutions, we extended Decision Service  \cite{agarwal2016making}, a simple interface for any developer 
to use RL. To learn more details about the implementation of our RL system and practical lessons, refer to \cite{karampatziakis2019lessons}.

\section{Conclusions and Future Work}
In this work, we describe two applications of CBs in the Microsoft virtual agent in the areas of intent disambiguation and contextual recommendations. We have demonstrated the efficacy of these solutions through online A/B experiments in the Microsoft virtual agent. The NLB solution for intent disambiguation provided 12.45\% improvement in the primary KPI of the Microsoft virtual agent, PRR. Our MAB models for contextual recommendations also increased PRR by 2.36\% in addition to increasing SHS and UE, while lowering EAS (escalations to a human agent). We have since deployed these solutions to production and impact millions of users each day. 

For future work, we are exploring how we can apply the insights from MABs and NLBs to multi-domain support bots as well as other products. We are also exploring real-world applications of episodic RL where it is important to 
do proper credit assignment. For this we
plan to rely on a reduction approach \cite{daume2018residual} which can operate 
well with the rest of our existing system.

\begin{acks}
We thank Paul Mineiro and John Langford 
for providing research advice for the CB scenario,
Eslam Kamal and the 
Microsoft Power Virtual Agents Team for their support with 
the Intent Disambiguation use case, Mary Buck, Sean Quigley, and the
Microsoft Digital Customer Support team for their help with both the 
Contextual Recommendation and Intent Disambiguation scenarios.
\end{acks}

\bibliographystyle{ACM-Reference-Format}
\bibliography{biblio}

\appendix

\end{document}